\documentclass[sigconf]{aamas}

\PassOptionsToPackage{numbers,sort&compress}{natbib}
\usepackage{balance} 
\usepackage[utf8]{inputenc} 
\usepackage[T1]{fontenc}    
\usepackage{hyperref}       
\usepackage{url}            
\usepackage{booktabs}       
\usepackage{amsfonts}       
\usepackage{nicefrac}       
\usepackage{microtype}      
\usepackage{xcolor}         
\usepackage{multirow}
\usepackage{makecell}
\usepackage{graphicx}
\usepackage{subcaption}     
\usepackage[ruled,vlined,linesnumbered]{algorithm2e}
\usepackage{amsmath}      
\usepackage{amsthm}       
\usepackage{bm}           

\usepackage{color}

\doi{HUOT2523}

\makeatletter
\gdef\@copyrightpermission{
  \begin{minipage}{0.2\columnwidth}
   \href{https://creativecommons.org/licenses/by/4.0/}{\includegraphics[width=0.90\textwidth]{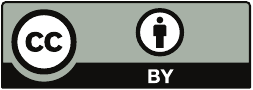}}
  \end{minipage}\hfill
  \begin{minipage}{0.8\columnwidth}
   \href{https://creativecommons.org/licenses/by/4.0/}{This work is licensed under a Creative Commons Attribution International 4.0 License.}
  \end{minipage}
  \vspace{5pt}
}
\makeatother

\setcopyright{ifaamas}
\acmConference[AAMAS '26]{Proc.\@ of the 25th International Conference
on Autonomous Agents and Multiagent Systems (AAMAS 2026)}{May 25 -- 29, 2026}
{Paphos, Cyprus}{C.~Amato, L.~Dennis, V.~Mascardi, J.~Thangarajah (eds.)}
\copyrightyear{2026}
\acmYear{2026}
\acmDOI{}
\acmPrice{}
\acmISBN{}

\title[]{
Pareto-guided Pipeline for Distilling Featherweight AI Agents\\
in Mobile MOBA Games
}

\author{Xionghui Yang}
\affiliation{
  \department{School of Computer Science}
  \institution{Peking University}
  \city{Beijing}
  \country{China}}
\email{yangxionghui@stu.pku.edu.cn}

\author{Bozhou Chen}
\affiliation{
  \department{School of Computer Science}
  \institution{Peking University}
  \city{Beijing}
  \country{China}}
\email{2301111899@stu.pku.edu.cn}

\author{Yunlong Lu}
\affiliation{
  \department{School of Computer Science}
  \institution{Peking University}
  \city{Beijing}
  \country{China}}
\email{luyunlong@pku.edu.cn}

\author{Yongyi Wang}
\affiliation{
  \department{School of Computer Science}
  \institution{Peking University}
  \city{Beijing}
  \country{China}}
\email{wangyongyi@pku.edu.cn}

\author{Lingfeng Li}
\affiliation{
  \department{School of Computer Science}
  \institution{Peking University}
  \city{Beijing}
  \country{China}}
\email{lingfengli@stu.pku.edu.cn}

\author{Lanxiao Huang}
\affiliation{
  \department{TiMi L1 Studio}
  \institution{Tencent}
  \city{Chengdu}
  \country{China}}
\email{jackiehuang@tencent.com}

\author{Lin Liu}
\affiliation{
  \department{TiMi L1 Studio}
  \institution{Tencent}
  \city{Chengdu}
  \country{China}}
\email{linliu@tencent.com}

\author{Wenjun Wang}
\affiliation{
  \department{TiMi L1 Studio}
  \institution{Tencent}
  \city{Chengdu}
  \country{China}}
\email{jamesonwang@tencent.com}

\author{Meng Meng}
\affiliation{
  \department{TiMi L1 Studio}
  \institution{Tencent}
  \city{Chengdu}
  \country{China}}
\email{promengmeng@tencent.com}

\author{Xia Lin}
\affiliation{
  \department{TiMi L1 Studio}
  \institution{Tencent}
  \city{Chengdu}
  \country{China}}
\email{hugolin@tencent.com}

\author{Wenxin Li}
\affiliation{
  \department{School of Computer Science}
  \institution{Peking University}
  \city{Beijing}
  \country{China}}
\email{lwx@pku.edu.cn}

\begin{abstract}
Recent advances in game AI have demonstrated the feasibility of training agents that surpass top-tier human professionals in complex environments such as Honor of Kings (HoK), a leading mobile multiplayer online battle arena (MOBA) game.
However, deploying such powerful agents on mobile devices remains a major challenge.
On one hand, the intricate multi-modal state representation and hierarchical action space of HoK demand large, sophisticated policy networks that are inherently difficult to compress into lightweight forms.
On the other hand, production deployment requires high-frequency inference under strict energy and latency constraints on mobile platform.
To the best of our knowledge, bridging large-scale game AI and practical on-device deployment has not been systematically studied.
In this work, we propose a Pareto optimality guided pipeline and design a high-efficiency student architecture search space tailored for mobile execution, 
enabling systematic exploration of the trade-off between performance and efficiency.
Experimental results demonstrate that the distilled model achieves remarkable efficiency, including an $12.4\times$ faster inference speed (under 0.5ms per frame) and a $15.6\times$ improvement in energy efficiency (under 0.5mAh per game), while retaining a 40.32\% win rate against the original teacher model.
\end{abstract}


\newcommand{\BibTeX}{\rm B\kern-.05em{\sc i\kern-.025em b}\kern-.08em\TeX}

\begin{document}


\pagestyle{fancy}
\fancyhead{}


\maketitle

\section{Introduction}
\label{sec: intro}

The development of powerful AI agents for complex environments such as mobile multiplayer online battle arena (MOBA) games marks a major milestone in artificial intelligence.
In particular, agents developed for Honor of Kings (HoK) have demonstrated the ability to defeat top-tier professional human players~\cite{ye2020towards, ye2020mastering, ye2020supervised}.
These state-of-the-art systems are typically implemented as large, deep neural networks with intricate, multi-branch architectures designed to process multi-modal state inputs and hierarchical action spaces~\cite{HoK3v3}.
Consequently, they demand substantial computational resources, often requiring hundreds of millions of floating-point operations (FLOPs) per inference.

A fundamental tension exists between the architectural complexity required for high performance and the stringent constraints of mobile platform—low latency, limited energy budget, and small memory footprint.
Directly deploying large models on-device typically leads to unacceptable inference latency and rapid battery drain, rendering these agents impractical for real-time mobile gaming.

Conventional model compression techniques, such as knowledge distillation (KD)~\cite{hinton2015distilling}, low-rank decomposition  (LRD)~\cite{sainath2013low}, quantization~\cite{hubara2018quantized}, and pruning~\cite{han2015learning}, offer only partial solution.
They struggle to cope with the unique characteristics of complex MOBA policy networks, which combine heterogeneous components (e.g., CNNs~\cite{lecun2002gradient, krizhevsky2012imagenet}, LSTMs~\cite{hochreiter1997long, gers2000learning}, and Transformers~\cite{vaswani2017attention}) under strong inter-module coupling through skip connections and feature fusion.
Applying generic compression uniformly across such architectures often results in severe performance degradation, as perturbations in one component propagate throughout the system.
More importantly, existing works lack a principled, reproducible pipeline for mobile deployment, leaving a significant methodological gap between research prototypes and deployable systems.

To address this gap, we reframe the deployment problem as a multi-objective optimization task. We propose a structured, Pareto-oriented engineering pipeline to approximate the Pareto-optimal frontier between agent performance and on-device efficiency. Our approach integrates a distillation-based architecture search pipeline specifically engineered for the MOBA domain. This end-to-end methodology transforms a complex teacher agent into a deployable mobile counterpart, embedding efficiency considerations at every design stage rather than treating them as an afterthought.

Our main contributions are as follows:
\begin{itemize}
    \item \textbf{Pareto-Guided Systematic Pipeline.} We introduce and formalize the problem of mobile deployment for large-scale multi-agent systems from a Pareto optimality~\cite{pareto2014manual} perspective, proposing an end-to-end engineering pipeline that systematically explores the trade-off between performance and efficiency.
    \item \textbf{Lightweight Student Architecture Design.} We design a high-efficiency student architecture optimized for policy distillation in mobile MOBA environments, achieving substantial computational and energy savings with only minor performance degradation.
    \item \textbf{Comprehensive Empirical Validation.} Through extensive experiments, we demonstrate that our method achieves an $12.4\times$ inference speedup and a $15.6\times$ improvement in energy efficiency while maintaining a competitive win rate against the teacher model. We further provide detailed ablations and design analyses, offering practical insights for future research on mobile game AI deployment.
\end{itemize}

The remainder of this paper is organized as follows: Section~\ref{sec: related work} reviews related work. Section~\ref{sec: preliminaries} formulates the problem and describes the HoK environment. Section~\ref{sec: method} details our proposed methodology. Section~\ref{sec: experiment} presents experimental results. Finally, Section~\ref{sec: conclusion} concludes the paper and discusses future research directions.


\section{Related Work}
\label{sec: related work}

Model compression has been a central topic in efficient deep learning, aiming to accelerate neural networks and reduce their size while maintaining competitive accuracy. Extensive research in both academia and industry has produced four major categories of methods: low-rank decomposition, pruning, quantization, and knowledge distillation.  
LRD methods compress networks by approximating the weight tensors of linear and convolutional layers through tensor decomposition, thereby reducing computational redundancy while preserving a degree of structural interpretability~\cite{jaderberg2014speeding, kim2015compression, balavzevic2019tucker, novikov2020tensor}. However, their rigid structural constraints often lead to accuracy degradation in complex, highly nonlinear models~\cite{jaderberg2014speeding, kim2015compression, idelbayev2020low}.  
Pruning methods, in contrast, selectively remove redundant neurons or connections to achieve sparsity. Structured pruning and automated strategies have enabled high compression rates with minimal accuracy loss~\cite{han2015learning, li2016pruning, he2018amc, frankle2018lottery, frantar2023sparsegpt}, though excessive pruning can destabilize training and impair generalization~\cite{frankle2018lottery, liu2018rethinking}.  
Quantization reduces the numerical precision of weights and activations, making models more hardware-friendly and improving inference efficiency~\cite{han2015deep, jacob2018quantization, hubara2018quantized, nagel2020up, dettmers2022gpt3}. Compared to LRD and pruning, quantization offers greater runtime acceleration but introduces numerical instability that must be mitigated through robust quantization-aware training~\cite{jacob2018quantization, wang2019haq, nagel2020up}.  
Finally, KD bridges the performance gap left by structural compression methods by transferring knowledge from a large teacher model to a compact student model~\cite{hinton2015distilling, romero2014fitnets, chen2017learning, zhang2021self}. KD preserves accuracy without imposing architectural constraints, yet its effectiveness critically depends on the alignment between teacher and student network designs.

KD has proven effective across domains such as computer vision and natural language processing~\cite{sanh2019distilbert, jiao2019tinybert, gu2023minillm, zhou2023distillspec, habib2024comprehensive, xu2024survey}. Its reinforcement learning counterpart, policy distillation (PD), extends this concept to compress deep reinforcement learning agents~\cite{rusu2015policy, czarnecki2019distilling, sun2019real, qu2022importance, yu2024online, spigler2024proximal, xu2024rldg}. However, most PD studies to date have been confined to relatively simple benchmarks such as Atari games~\cite{bellemare2013arcade}, which lack the multi-agent coordination, continuous dynamics, and high-dimensional inputs present in MOBA environments.  
In contrast, the HoK environment poses a substantially greater challenge. Unlike Atari, which uses low-dimensional pixel inputs and a small discrete action space (18 actions), HoK requires real-time reasoning over multi-modal state representations, a hierarchical action space with over 22 million possible actions, and partial observability across multiple agents~\cite{HoK3v3}. These properties make HoK an ideal testbed for studying large-scale policy compression under realistic and demanding conditions.

Despite progress in model and policy compression, most existing approaches overlook the stringent requirements of mobile environments, where policies must support high-frequency, low-latency, and energy-efficient inference under severe hardware constraints. These challenges are particularly acute in mobile MOBA games, which require consistent responsiveness and prolonged device operation.  
Critically, current compression approaches provide only partial solutions and lack a principled, reproducible pipeline for real-world mobile deployment. This gap leaves a disconnect between research prototypes and deployable, resource-efficient systems. Consequently, distilling large-scale policy models from complex environments like HoK into computationally and energy-efficient student agents remains a challenging and underexplored problem.


\section{Preliminaries}
\label{sec: preliminaries}

This section formalizes the core deployment challenge and introduces the experimental environment. We first frame the problem using multi-objective optimization and then describe the HoK environment that serves as our testbed.

\subsection{Problem Formulation}
\label{ssec: problem formulation}

The deployment of complex game AI agents onto mobile devices presents a fundamental conflict: the \emph{computational demand} for high-performance agents versus the stringent \emph{resource constraints} of mobile platforms. We formalize this challenge as a multi-objective optimization problem, where the goal is to find an optimal trade-off between competing objectives.

Consider a design point \(d\) within a vast design space \(D\), encompassing all possible model architectures and their configurations. We evaluate \(d\) along two primary axes:

\begin{itemize}
    \item \textbf{Performance (\(P(d)\))}: The efficacy of the agent, quantified as its win rate against a benchmark.
    \item \textbf{Efficiency (\(E(d)\))}: The on-device resource consumption, which we decompose into:
    \begin{itemize}
        \item Inference Latency (\(L(d)\)): Time per decision.
        \item Energy Consumption (\(B(d)\)): Power used per decision.
        \item Peak Memory Usage (\(M(d)\)): Maximum RAM during inference.
        \item Model Size (\(S(d)\)): Storage footprint.
    \end{itemize}
\end{itemize}

The ideal solution would simultaneously maximize \(P(d)\) while minimizing all efficiency metrics. However, these objectives are inherently competing. This tension leads us to adopt the concept of \textbf{Pareto optimality} to define the best possible compromises.

\begin{definition}[Pareto Dominance]
A point \(d_i\) \textit{Pareto dominates} another point \(d_j\) (denoted \(d_i \prec d_j\)) if and only if \(d_i\) is strictly better in at least one objective and no worse in all others. Formally, for the primary objectives \((P, L, B, M, S)\):
\begin{align}
\begin{split}
[P(d_i) \geq P(d_j)] \land [L(d_i) \leq L(d_j)] \land [B(d_i) \leq B(d_j)]\\
\land [M(d_i) \leq M(d_j)] \land [S(d_i) \leq S(d_j)]
\end{split}
\end{align}
with at least one inequality being strict.
\end{definition}

\begin{definition}[Pareto Optimality and Pareto Frontier]
A point \(d^*\) is \textit{Pareto-optimal} (non-dominated) if no other point in \(D\) dominates it. The set of all Pareto-optimal points forms the \textit{Pareto frontier}, representing the best achievable trade-offs.
\end{definition}

Given the intractability of finding the global Pareto frontier for \(D\), our practical goal is to procedurally generate a restricted design subspace \(D' \subset D\) and identify one or more design points \(\hat{d} \in D'\) that are non-dominated and lie on or near the \emph{empirical Pareto frontier} of \(D'\).

\begin{figure*}[h]
    \centering
    \includegraphics[width=\linewidth]{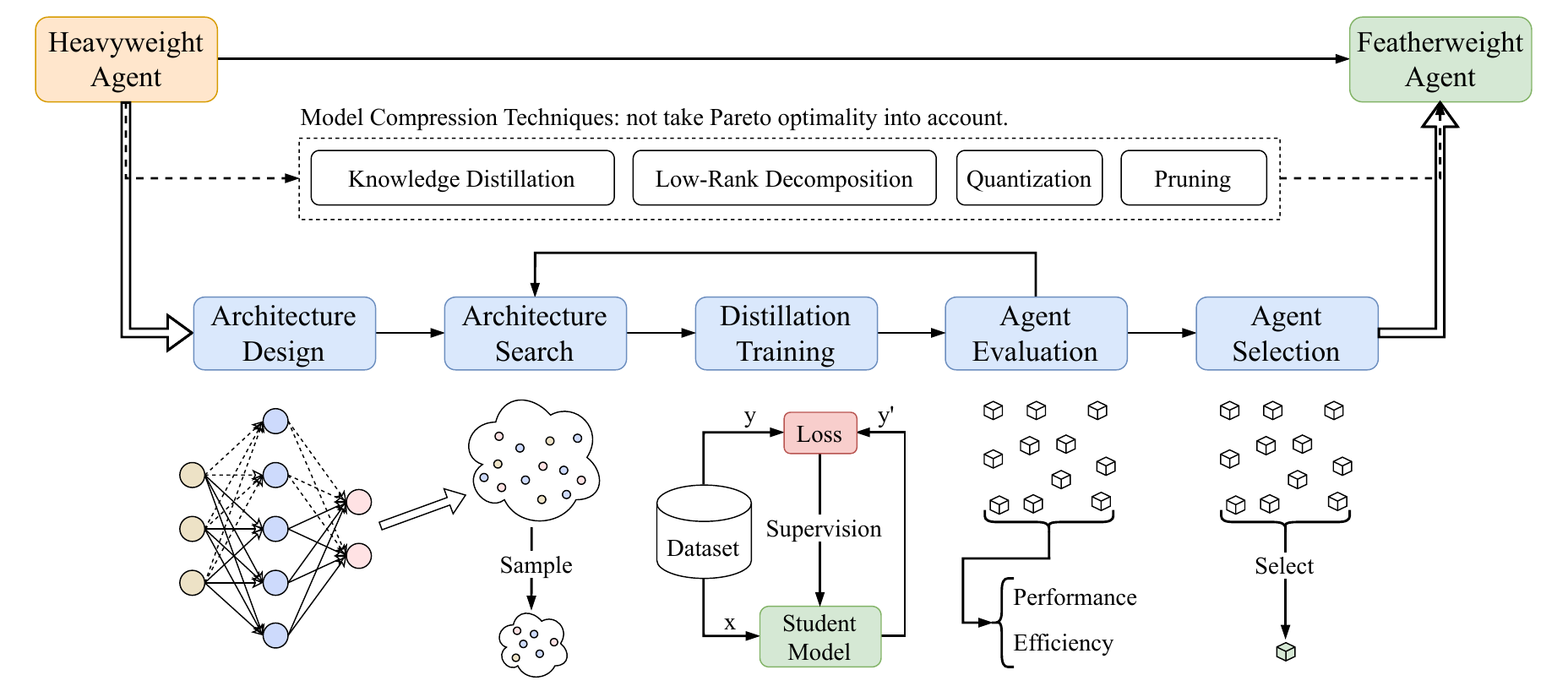}
    \caption{Overview of the proposed Pareto optimality driven distillation pipeline. The process integrates architecture design, automated search, distillation training, evaluation, and final selection, forming an end-to-end framework that jointly optimizes model performance and efficiency for mobile deployment.}
    \label{fig: pipeline}
\end{figure*}

\subsection{HoK Environment}
\label{ssec: hok environment}

We employ the HoK 3v3 game mode as our experimental testbed, a leading mobile MOBA game that presents a challenging benchmark for multi-agent decision-making under real-time constraints. In this environment, two teams of three agents (heroes) compete to destroy the opponent's base, requiring sophisticated coordination and rapid, precise micromanagement.

The environment is characterized by two core elements that define the complexity of the learning and deployment problem:

\textbf{High-Dimensional, Multi-Modal Observation Space.}
The observation space is represented by a 13,758-dimensional vector that comprehensively encodes the game state at each time step. This representation is decomposed into three hero-specific observations, each of 4,586 dimensions, capturing:
\begin{itemize}
    \item Individual hero attributes (e.g., health, skill cooldowns, position),
    \item Environmental obstacles and terrain features,
    \item The dynamic status of all game units (e.g., creeps, turrets, monsters).
\end{itemize}
As detailed in Table~\ref{tab: obs space} in Appendix~\ref{app sec: hok env}, these features are systematically organized into seven distinct categories, providing agents with a holistic perception necessary for strategic decision-making.

\textbf{Hierarchical Action Space with Validity Constraints.}
The action space adopts a two-level hierarchical structure (as illustrated in Figure~\ref{fig: two level concept} and detailed in Tables~\ref{tab: action space} and \ref{tab: detail action space}  in Appendix~\ref{app sec: hok env}) to enable fine-grained control:
\begin{itemize}
    \item \textbf{Level 1 (Action Selection):} Chooses a behavior type from 13 discrete actions, including 2 No-operations, Move, Normal Attack, and 9 hero-specific skills.
    \item \textbf{Level 2 (Parameter Specification):} Determines the parameters for the chosen action through three sub-mechanisms:
    \begin{itemize}
        \item Directional Control: 25 discrete movement directions.
        \item Positional Control: 42 possible skill offsets along the x-axis and z-axis.
        \item Target Selection: 7 target types with up to 39 possible targets.
    \end{itemize}
\end{itemize}
To ensure that actions are contextually valid, the environment incorporates \textit{legal action masks} and \textit{sub-action masks}~\cite{ye2020mastering, ye2020towards}, which dynamically restrict the available choices to those permissible in the current game state. These mechanisms, detailed in Appendix~\ref{app ssec: legal action mask} and~\ref{app ssec: sub action mask}, are critical for stabilizing training and ensuring the practicality of the learned policies.

The combination of a high-dimensional, multi-modal state space and a complex, hierarchical action space makes HoK 3v3 an ideal and demanding platform for studying the performance-efficiency trade-offs in mobile AI deployment.

\section{Methodology}
\label{sec: method}

\begin{figure*}[h]
    \centering
    \includegraphics[width=1\linewidth]{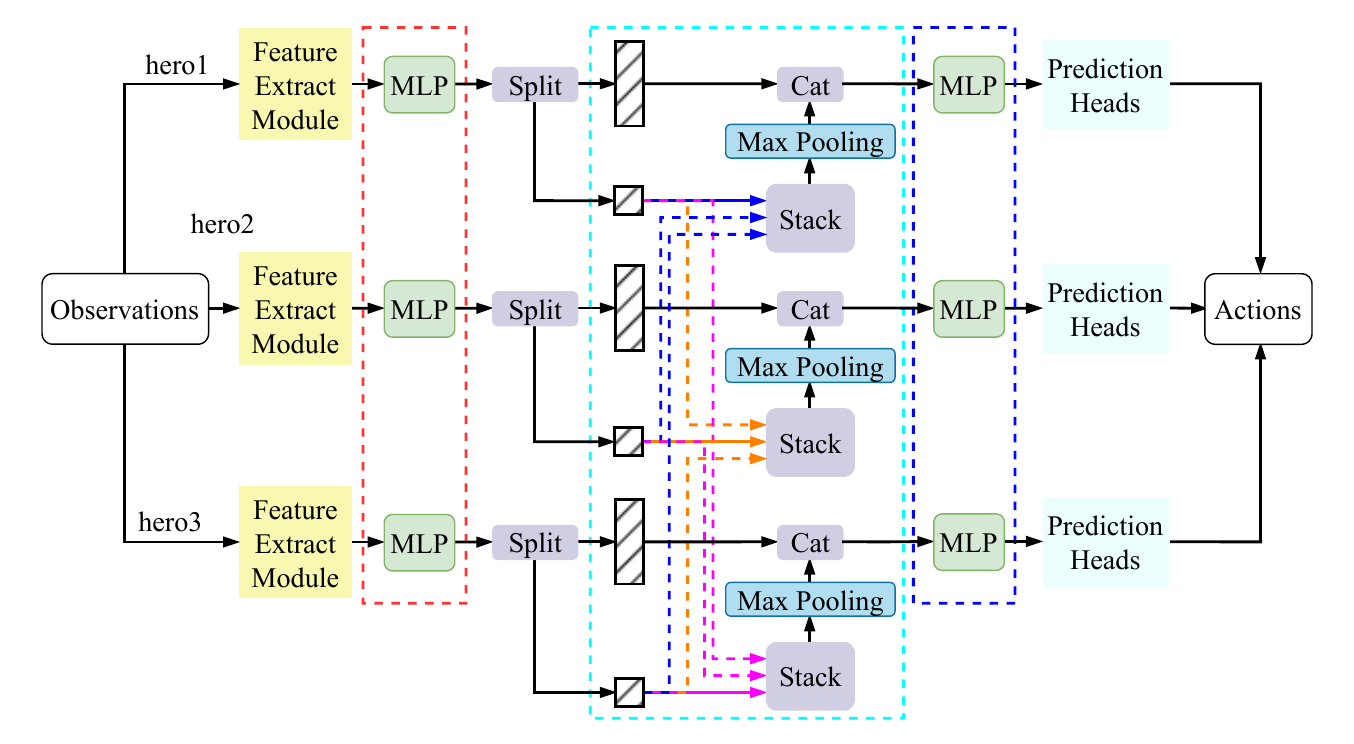}
    \caption{Overview of the featherweight student architecture. The design streamlines the teacher model by removing the attention-based components and the LSTM module, and adopting lightweight MLP structures. The red and blue dashed boxes indicates the simplified feature fusion module, while the cyan dashed boxes highlights the triplet max-fusion gate that enables team-level cooperation. This architecture efficiently processes multi-modal inputs and produces hierarchical action distributions.}
    \label{fig: student architecture}
\end{figure*}

\subsection{Overview}
\label{ssec: overview}

Following the problem formulation in Section~\ref{sec: preliminaries}, which frames the deployment challenge as a multi-objective optimization targeting Pareto optimality, we introduce a systematic multi-stage pipeline. Our overarching goal is to transform a computationally intensive teacher agent into a featherweight student agent suitable for mobile deployment while optimally navigating the performance-efficiency trade-off.

The proposed pipeline, depicted in Figure~\ref{fig: pipeline}, comprises five stages:
\begin{enumerate}
    \item \textbf{Architecture Design}: Profiling the teacher network to identify computational bottlenecks and design a mobile-efficient student blueprint with tunable hyperparameters.
    \item \textbf{Architecture Search}: Conducting a systematic search over the designed space, distilling each candidate, and evaluating its performance-efficiency profile to progressively construct the Pareto frontier.
    \item \textbf{Distillation Training}: Transferring policy knowledge from the teacher to the candidate students via knowledge distillation.
    \item \textbf{Agent Evaluation}: Assessing candidates using key performance (e.g., win rate) and efficiency (e.g., latency, energy) metrics.
    \item \textbf{Agent Selection}: Selecting the optimal agent from the empirical Pareto frontier based on predefined deployment constraints.
\end{enumerate}

\subsection{Architecture Design}
\label{ssec: architecture design}

We commence with a rigorous profiling of the teacher model (see architecture details in Appendix~\ref{app sec: teacher architecture}) to identify computational bottlenecks. The profiling results, summarized in Table~\ref{tab:teacher_profile}, provide a quantitative breakdown of the computational cost distribution across different modules.

\begin{table}[h]
\centering
\caption{Computational profile of the teacher model. The encoder and LSTM modules are identified as the primary bottlenecks.}
\label{tab:teacher_profile}
\begin{tabular*}{\linewidth}{@{}@{\extracolsep{\fill}}lrrrr@{}}
\toprule
\textbf{Module} & \textbf{FLOPs (M)} & \textbf{Params (M)} & \textbf{FLOPs(\%)} & \textbf{Params(\%)} \\
\midrule
All & 681.84 & 16.43 & 100.00 & 100.00 \\
\textbf{Encoder} & \textbf{560.92} & \textbf{4.83} & \textbf{82.27} & \textbf{29.39} \\
\textbf{LSTM} & \textbf{8.40} & \textbf{8.40} & \textbf{1.23} & \textbf{51.11} \\
CNN & 5.06 & 0.004 & 0.74 & 0.02 \\
Others & 107.45 & 3.20 & 15.76 & 19.47 \\
\bottomrule
\end{tabular*}
\end{table}

Our analysis reveals that the encoder and LSTM modules collectively account for 83.49\% of the computational cost and 80.50\% of the model parameters, despite the LSTM contributing negligible FLOPs. This stark imbalance establishes these two modules as the primary bottlenecks and the most promising targets for architectural optimization.

Guided by this profiling, we design a featherweight and effective student architecture. As shown in Figure~\ref{fig: student architecture}, the architecture is tailored to address the challenges of HoK's multi-modal inputs and hierarchical action space while preserving the functional integrity of the teacher model. Its core components are structured as follows:

\begin{enumerate}
    \item \textbf{Feature Extraction Module}: Processes each hero's multi-modal observations in parallel using a combination of CNNs and MLPs.
    \item \textbf{Feature Fusion Module}:
    In the teacher model, the Encoder and LSTM handle cross-modal and cross-temporal fusion, respectively.  In our student model, both functions are implemented as MLPs (see the red and blue dashed boxes in Figure~\ref{fig: student architecture}).
    \item \textbf{Triplet Max-Fusion Gate}: Facilitates cooperation among heroes by sharing and max-pooling public embeddings across the team (indicated by the blue dotted line in Figure~\ref{fig: student architecture}).
    \item \textbf{Prediction Heads}: Comprise specialized MLPs that output distributions for the hierarchical action space.
\end{enumerate}


To further refine the student model, we parameterize the architecture using a set of tunable hyperparameters summarized in Table~\ref{tab: search_space} in Appendix~\ref{app sec: search space}, which serves as the foundation for automated architecture search in the subsequent stage.

\subsection{Architecture Search}
\label{ssec: architecture search}

Building upon the macro-architecture defined in Section~\ref{ssec: architecture design}, we conduct a systematic search within our architectural space to identify high-performing candidate agents. To manage the search complexity and ensure a balanced distribution of candidates, we employ a constrained sampling strategy. The space is decomposed into seven core building blocks (Table~\ref{tab: search_space} in Appendix~\ref{app sec: search space}), each parameterized by its layer count and per-layer feature dimension $n_i$. Candidate architectures are sampled from these parameter ranges, subject to a global compute budget of $[1\%, 20\%]$ of the teacher's FLOPs.

We discretize the FLOPs range into 20 uniform intervals and sample candidate models within each interval. This structured approach ensures efficient and representative coverage of the performance-efficiency design space, providing a diverse population for subsequent Pareto-optimal selection.

Our search procedure is outlined in Algorithm~\ref{alg:pareto_nas_distill}. For each architecture $q \in \mathcal{S}$, we obtain its distilled weights $\theta_q$ and an evaluation vector $\mathbf{m}_q \in \mathbb{R}^d$ (e.g., win rate, latency, energy). The accumulated results after round t form the dataset $\mathcal{D}_t = \{(q, \mathbf{m}_q)\}$, from which we derive the Pareto frontier $\mathcal{P}_t$.

\begin{algorithm}[t]
\caption{Pareto-Guided Architecture Search with Distillation}
\label{alg:pareto_nas_distill}
\DontPrintSemicolon
\SetKwInput{KwIn}{Input}
\SetKwInput{KwOut}{Output}

\KwIn{Search space $\mathcal{S}$; teacher model $\pi_T$; architecture sampler $\mathcal{Q}(\cdot\mid\mathcal{S})$; batch size $k$; max iterations $R$; threshold $\varepsilon>0$.}
\KwOut{Pareto-optimal architecture set $\mathcal{A}^{\star}$.}

\BlankLine
$\mathcal{D}_0 \leftarrow \varnothing$, $\mathcal{P}_0 \leftarrow \varnothing$.\;

\For{$t \leftarrow 1$ \KwTo $R$}{
  \tcp{(a) sample architectures}
  $\mathcal{A}_t \leftarrow \left\{\,q\,|\, q \sim \mathcal{Q}(\cdot\mid\mathcal{S}) \,\right\}$\;

  \ForEach{$q \in \mathcal{A}_t$}{
    \tcp{(b) distillation training (teacher $\pi_{T}$)}
    Train student $\pi_q(\cdot|s;\theta)$ by minimizing
    $\mathcal{L}(\theta) = \text{KL}\left( \pi_T(\cdot|s) \ \middle\| \ \pi_q(\cdot|s;\theta) \right),$ obtain $\theta_q$.\;

    \tcp{(c) evaluation}
    Measure $\mathbf{m}_q$ on validation set and system metrics (FLOPs/accuracy/win rate/latency/energy, etc.).\;
    $\mathcal{D}_t \leftarrow \mathcal{D}_{t-1} \cup \{(q,\mathbf{m}_q)\}$\;
  }

  \tcp{(d) compute new Pareto frontier}
  $\mathcal{P}_t \leftarrow \{\,\mathbf{u}\in\{\mathbf{m}_q\}\ \mid\ \nexists\,\mathbf{v}\text{ s.t. }\mathbf{v}\preceq \mathbf{u}\ \&\ \mathbf{v}\neq \mathbf{u}\,\}$\;

  \tcp{(e) stopping by frontier discrepancy}
  \emph{(Example metric: symmetric Chamfer distance in normalized objective space)}
  \[
  \Delta_t \leftarrow 
  \frac{1}{|\mathcal{P}_t|}\!\sum_{\mathbf{u}\in\mathcal{P}_t}\min_{\mathbf{v}\in\mathcal{P}_{t-1}}\!\|\mathbf{u}-\mathbf{v}\|_2
  \;+\;
  \frac{1}{|\mathcal{P}_{t-1}|}\!\sum_{\mathbf{v}\in\mathcal{P}_{t-1}}\min_{\mathbf{u}\in\mathcal{P}_t}\!\|\mathbf{v}-\mathbf{u}\|_2
  \]
  \If{$\Delta_t < \varepsilon$}{
    $t^\star \leftarrow t$; \textbf{break}\;
  }
}

$\mathcal{A}^{\star} \leftarrow \{\, q \mid (q,\mathbf{m}_q)\in\mathcal{D}_{t^\star},\ \mathbf{m}_q \in \mathcal{P}_{t^\star} \,\}$.\;

\end{algorithm}

\subsection{Distillation Training}
\label{ssec: distillation}

Knowledge transfer from the teacher to the student is formalized as a policy distillation task. The core objective is to align the student’s action distribution with the teacher’s policy, enabling the lightweight model to retain expert-level decision quality while achieving significant efficiency gains.

Given an input state s sampled from replay dataset $\mathcal{B}$, let $\hat{z}$ and z be the raw logits of the teacher and student, respectively. The temperature-scaled probability distributions are:
\begin{equation}  
\label{equ: pd_distri}  
\hat{p}_i = \frac{\exp(\hat{z}_i / \tau)}{\sum_j \exp(\hat{z}_j / \tau)}, \quad p_i = \frac{\exp(z_i / \tau)}{\sum_j \exp(z_j / \tau)},
\end{equation} 
where $\tau$ controls the smoothness of the output distribution and aids knowledge transfer.

The student policy $\pi_{\theta}$ is optimized to minimize the divergence from the teacher policy $\pi_T$ over the dataset:
\begin{equation}  
\label{equ: pd} 
\min_{\theta} \mathbb{E}_{s \sim \mathcal{B}} \left[ \mathcal{D}\left( \pi_T(\cdot|s) \ \middle\| \ \pi_{\theta}(\cdot|s) \right) \right],
\end{equation}

For the hierarchical action space of HoK with $\mathcal{A}
 = 5$ sub-distributions, we define the loss as the average KL divergence across all sub-actions:
\begin{equation}
\label{equ: pd_loss}
\mathcal{L}(\theta) = \frac{1}{|\mathcal{A}|} \sum_{k=1}^{|\mathcal{A}|} \mathbb{E}_{s \sim \mathcal{B}} \left[ \text{KL}\left( \pi_T^{(k)}(\cdot|s) \ \middle\| \ \pi_{\theta}^{(k)}(\cdot|s) \right) \right],
\end{equation}
where distributions are softened with $\tau$ as in Equation~\ref{equ: pd_distri}.

This distillation process is applied to each candidate architecture sampled during the search stage, training a diverse population of models for the subsequent Pareto-optimal selection.



\subsection{Agent Evaluation and Selection}
\label{ssec: evaluation and selection}

In the evaluation phase, Each candidate model is evaluated across the dual objectives of \textbf{performance} (competitive ability against the teacher model) and \textbf{efficiency} (on-device resource consumption), using the metrics described in Section~\ref{ssec: experimental setup}, encompassing win rate, inference latency, energy consumption, peak memory usage, and model size.

The final agent is selected from the empirical Pareto frontier according to the principle of Pareto optimality: we choose the non-dominated candidate that satisfies predefined mobile deployment constraints (e.g., win rate  > 40\% , latency  < 0.5 \text{ms} ). This constraint-driven approach ensures the selected agent optimally balances performance and efficiency for practical on-device deployment.



\section{Experiments}
\label{sec: experiment}

\subsection{Experimental Setup}
\label{ssec: experimental setup}
\textbf{Environment.}
Our experiments are conducted within the HoK 3v3 game mode on the "Changping Attack and Defense" map. To ensure a robust and comprehensive evaluation, we employ a fixed set of eight distinct hero compositions (detailed in Appendix~\ref{app sec: hero compositions}). These compositions are systematically generated from a pool of six popular heroes (e.g., Zhao Yun, Li Yuanfang, Zhuge Liang), covering a variety of strategic roles and team synergies. This approach mitigates potential evaluation bias from testing on a single, favorable composition, thereby providing a more reliable assessment of agent performance across diverse in-game scenarios.

\textbf{Dataset.}
The dataset for knowledge distillation is constructed through self-play between two identical instances of the teacher model, generating a corpus of 2.02 million state-action transitions. This dataset is partitioned into 2 million samples for training and 20,000 for validation.

\textbf{Hyper-parameters Configuration.}
The hyperparameter configuration in distillation training are described in Table~\ref{tab:hyper_parameter} in Appendix~\ref{app sec: hyper_parameters}).

\textbf{Baseline Methods.}
To rigorously evaluate our approach, we compare it against several representative baselines spanning different compression paradigms:
(1) \textbf{Teacher Model}: The original, uncompressed model serves as the performance upper bound, representing the maximum achievable capability before any compression is applied.
(2) \textbf{LRD}: Applies low-rank matrix factorization to the weight tensors of the teacher model, approximating the original parameters with lower-rank representations. Parameter count and computational complexity are reduced while the fundamental structure of the original network is preserved as much as possible.
(3) \textbf{Quantization}: Reduces the numerical precision of weights and activations from 32-bit floating point to 8-bit integers (INT8) through post-training quantization. This technique primarily decreases model size and improves inference speed on supported hardware, though potential precision loss may degrade policy quality.
(4) \textbf{Pruning}: Removes connections with the smallest magnitudes based on a global threshold via structured weight pruning. Sparsity in the weight matrices is increased, leading to reductions in parameter count and computational cost, albeit with possible loss of structural integrity and performance.
(5) \textbf{Linear Model}: An extremely simple architecture (a MLP) serves as the efficiency upper bound, representing the theoretical maximum efficiency achievable while typically exhibiting negligible performance.

\textbf{Metrics.}
Evaluation is conducted from two primary perspectives to holistically assess the agent's quality:
(1) \textbf{Performance (\(P(d)\))}: The primary metric is the \textit{Win Rate} against the original teacher model, calculated as the average win rate over three independent runs of 1,000 games each to ensure statistical significance.
(2) \textbf{Efficiency (\(E(d)\))}: Metrics are measured on mobile platforms and include:
\textit{Inference Latency (\(L(d)\))}: Time required for a single decision. This is considered the most critical efficiency metric. 
\textit{Energy Consumption (\(B(d)\))}: Power usage per game, it is the total power usage of 5,000 inferences. Generally speaking, the total number of frames in a game is less than 5000.
\textit{Peak Memory Usage (\(M(d)\))}: Maximum RAM allocated during inference.
\textit{Model Size (\(S(d)\))}: Storage footprint of the deployed model.
To ensure reliable and reproducible measurements of on-device efficiency, all metrics are evaluated on a fixed set of 5,000 frames randomly sampled from the validation set. This controlled approach mitigates the performance fluctuations inherent in live game environments. Each measurement was independently repeated three times, and the mean and standard deviation were reported.

\textbf{Hardware Configuration.}
All model training and distillation procedures were performed on a high-performance computing cluster consisting of four servers. Each server was equipped with a 112-core Intel Xeon Gold 6348 CPU (2.60 GHz), 378 GB of RAM, and eight NVIDIA GeForce RTX 3090 GPUs.
Mobile deployment and on-device efficiency evaluation were conducted on an iQOO 12 smartphone, featuring a Qualcomm Snapdragon 8 Gen 3 system-on-chip with an integrated Adreno GPU and Hexagon NPU, along with 16 GB of RAM. 

\textbf{Wall-Clock Time.}
A complete run of the proposed pipeline, including architecture search, distillation, and evaluation, completes within approximately 10-15 days on the hardware setup described above.

\subsection{The Performance of Proposed Pipeline}

This section presents a macro-level evaluation to answer the fundamental question: does our proposed pipeline achieve a superior trade-off between agent performance and on-device efficiency? We compare our final Featherweight Agent (FA) against the original teacher model and all baseline methods across the comprehensive set of metrics defined in Section~\ref{ssec: problem formulation}.

\textbf{Macro-Level Results.} The comprehensive benchmarking results are summarized in Table~\ref{tab: performance}. Our FA achieves a competitive win rate of \textbf{40.32\%}, 
well preserving the decision-making ability of the teacher model. Crucially, this performance is attained while achieving a dramatic reduction in resource consumption. Compared to the teacher model, FA reduces inference latency by 91.8\% (from 5.41ms to 0.44ms) and energy consumption by 93.6\% (from 7.62mAh to 0.49mAh). Furthermore, while Low-Rank Decomposition and Quantization maintain reasonable performance (\textasciitilde40\%), their efficiency gains are modest, achieving less than 50\% latency reduction. In contrast, Pruning achieves high efficiency but suffers a catastrophic performance collapse. Our FA, by fundamentally re-architecting the model to be mobile-native, achieving both high performance and superior efficiency. 

\begin{figure}[h]
    \centering
    \includegraphics[width=1\linewidth]{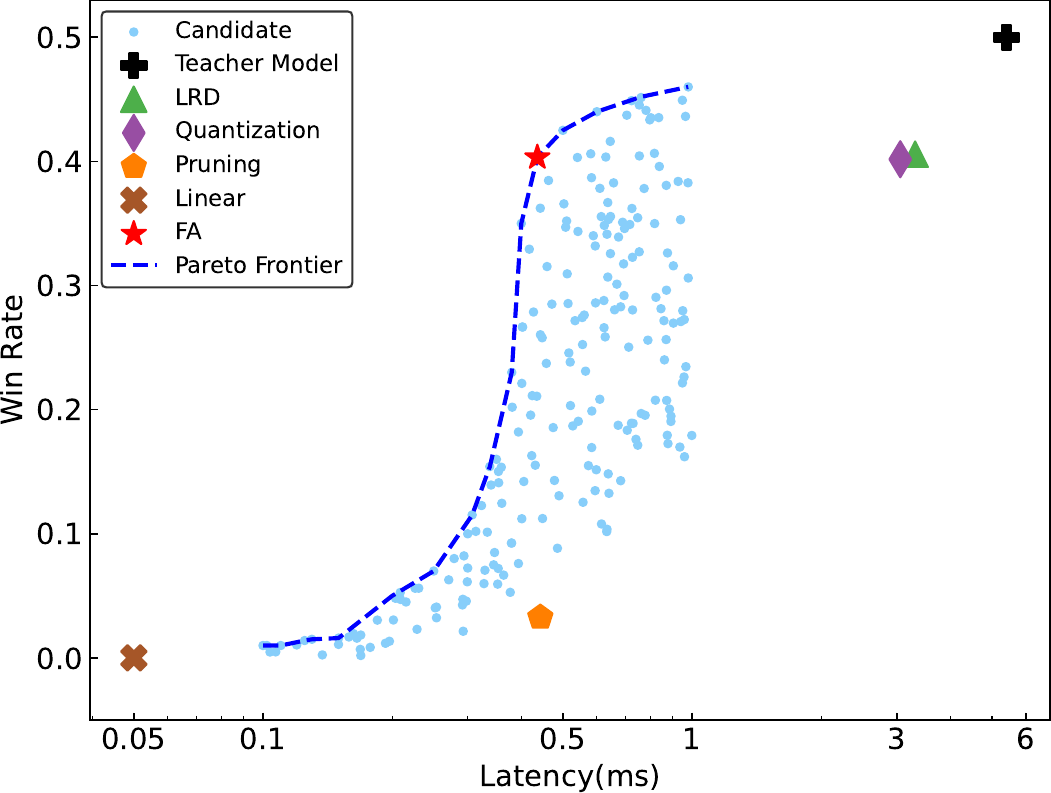}
    \caption{
    The empirical Pareto frontier characterizing the performance-efficiency trade-off. The frontier is derived from the systematic assessment of all candidate agents, illustrating the set of non-dominated solutions where no improvement can be made in one objective without deteriorating the other.
    }
    \label{fig: pareto}
\end{figure}

\textbf{Pareto Frontier Analysis.}
The performance-efficiency trade-off is visually articulated in Figure~\ref{fig: pareto}, which plots the empirical Pareto frontier. Each point represents a candidate model's win rate versus its inference latency. The plot clearly shows that our FA resides on the estimated Pareto frontier, demonstrating that it is a non-dominated solution. In contrast, the standard compression baselines lie deep within the interior of the plot, indicating they are strictly inferior in the trade-off space (i.e., other points offer better or equal performance with lower latency). This provides compelling visual evidence that our approach successfully identifies a superior operating point.

\begin{table*}[ht]
\caption{Comprehensive benchmarking results comparing the performance and efficiency of the teacher model, standard compression baselines, and our Featherweight Agent. Results demonstrate that FA achieves a superior trade-off, dominating other methods by simultaneously maintaining competitive performance (40.32\%win rate) while achieving dramatic efficiency improvements ($12.4\times$ faster inference speed and $15.6\times$ improvement in energy efficiency than the teacher). Metrics are reported as mean values with standard deviations in parentheses.}
\label{tab: performance}
\centering
\begin{tabular*}{\linewidth}{@{}@{\extracolsep{\fill}}lrrrrrrr@{}}
    \toprule  
    \textbf{Method}&\textbf{FLOPs(M)}&\textbf{Parameters(M)}&\textbf{Win Rate(\%)}&\textbf{Time(ms)}&\textbf{Energy(mAh)}&\textbf{Memory(MB)}&\textbf{Size(MB)}\\
    \midrule
    Teacher Model &681.84&16.43&49.98 ($\pm$4.59)&5.411 ($\pm$0.188)&7.62 ($\pm$0.27)&230.35 ($\pm$1.94)&32.7\\
    LRD &412.36&11.56&40.58 ($\pm$1.39)&3.312 ($\pm$0.104) &4.56 ($\pm$0.18)&181.22 ($\pm$0.98)&19.05\\
    Quantization &380.18&10.13&40.17 ($\pm$1.18)&3.058 ($\pm$0.096) &4.21 ($\pm$0.13)&176.39 ($\pm$0.93)&14.23\\
    Pruning &45.22&6.22&3.27 ($\pm$0.14)&0.443 ($\pm$0.005) &0.52 ($\pm$0.01)&156.62 ($\pm$0.17)&12.36\\
    Linear & 4.43 & 2.22 & 0.00 ($\pm$0.00) & 0.05 ($\pm$0.001) & 0.06 ($\pm$0.01) & 79.63 ($\pm$0.06) & 3.63 \\
    \textbf{FA (ours)} &\textbf{45.74}&\textbf{5.96}&\textbf{40.32 ($\pm$1.03)}&\textbf{0.436 ($\pm$0.006)}&\textbf{0.49 ($\pm$0.01)}&\textbf{152.71 ($\pm$0.19)}&\textbf{9.89}\\
    \bottomrule
\end{tabular*}
\end{table*}

\subsection{Comparison with Standard Compression Baselines}

This section validates the superiority of our pipeline by contrasting it against standard, architecture-preserving compression techniques: Low-Rank Decomposition, Quantization, and Pruning. As quantified in Table~\ref{tab: performance} and visually articulated in Figure~\ref{fig: pareto}, models compressed by these standard techniques reside far from the empirical Pareto frontier. While they reduce model size and latency, they incur a severe performance penalty. For instance, the pruned model's win rate collapses to 3.27\%, rendering it practically useless, despite its low latency (0.44ms). LRD and Quantization maintain \textasciitilde40\% win rates but achieve only modest latency reductions (\textasciitilde3ms).

The failure of each method stems from fundamental limitations when applied to complex policy networks: Pruning causes irreversible structural damage by removing critical pathways in the highly non-linear policy network, Quantization leads to information degradation as aggressive precision loss (e.g., INT8) disrupts the delicate policy function, Low-Rank Decomposition suffers from representational inadequacy, as low-rank constraints poorly approximate the full-rank functions needed for sophisticated policies.


In summary, while standard compression techniques are inherently architecture-preserving, merely compressing an existing, inefficient structure, our approach is fundamentally architecture-agnostic. By undertaking a ground-up, mobile-native redesign based on the teacher model, our Featherweight Agent successfully preserves the core decision-making capabilities while achieving superior efficiency. These results underscore a critical insight: when the source architecture is inherently ill-suited for the target platform, post-hoc compression is fundamentally limited in its ability to yield an optimal solution. This work demonstrates that a mobile-native, ground-up architectural redesign is a substantially more effective strategy.

\subsection{Ablation Study}
\label{ssec: ablation study}

We conduct a comprehensive set of ablation studies to rigorously validate the necessity and optimality of each key design choice in our Featherweight Agent. The results, summarized in Table~\ref{tab: ablation_results}, are analyzed to demonstrate how each alteration leads to a strictly inferior trade-off.

\begin{table}[h]
    \centering
    \caption{Ablation study results. Key metrics include win rate against the teacher, inference latency, and energy consumption.}
    \label{tab: ablation_results}
    \begin{tabular}{@{}lrcc@{}}
        \toprule
        Model Variant & Win Rate (\%) & Latency (ms) & Energy (mAh) \\
        \midrule
        Teacher Model & 49.98 ($\pm$4.59) & 5.411 ($\pm$0.188) & 7.62 ($\pm$0.27) \\
        \textbf{FA (ours)} & \textbf{40.32 ($\pm$1.03)} & \textbf{0.436 ($\pm$0.006)} & \textbf{0.49 ($\pm$0.01)} \\
        \hline
        FA (w/ LSTM) & 40.67 ($\pm$0.91) & 0.578 ($\pm$0.012) & 0.72 ($\pm$0.02) \\
        FA (w/ Encoder) & 45.21 ($\pm$1.64) & 4.656 ($\pm$0.128) & 6.55 ($\pm$0.21) \\
        FA (w/ Large) & 44.00($\pm$1.27) & 0.601 ($\pm$0.007) & 0.68 ($\pm$0.02) \\
        FA (w/ Small) & 23.02($\pm$0.98) & 0.380 ($\pm$0.005) & 0.43 ($\pm$0.01) \\
        \bottomrule
    \end{tabular}
\end{table}

\subsubsection{Impact of Different Network Modules}
To validate our architecture efficiency, we reintroduced the original complex modules into the FA. Reinstating the LSTM (FA w/ LSTM) increases latency by 32.6\% (0.436ms to 0.578ms) and energy consumption by 46.9\%, with no meaningful win rate improvement. Reinstating the encoder (FA w/ Encoder) has a more severe impact, increasing latency by 10.7$\times$ and energy consumption by 13.4$\times$, despite a modest win rate increase to 45.21\%.
These results demonstrate the reasonablility of our architecture design. The original complex modules are indeed computational bottlenecks, and replacing them with simpler alternatives is essential for mobile deployment.

\subsubsection{Effect of Model Scale}

An ablation study on the overall model scale validates the optimality of our selected architecture (FA). The results, described in Table~\ref{tab: ablation_results}, 
demonstrate the performance-efficiency trade-off in the vicinity of the Pareto frontier. 

Scaling up the model (FA w/ Large) yields a negligible performance gain of only 3.68\% in win rate, but incurs a substantial efficiency cost, with a 37.8\% increase in latency. Conversely, scaling down the model (FA w/ Small) achieves a minimal latency reduction of 12.8\% at the expense of a significant 42.9\% performance penalty.

Our analysis reveals that the chosen FA operates near a knee point on the empirical Pareto frontier. At this region, the trade-off curve exhibits a characteristic sharp bend: reducing the model scale beyond this point results in a disproportionate drop in performance for a minimal efficiency gain, while increasing the model scale leads to diminishing returns in performance accompanied by a steep efficiency loss. This confirms that FA represents a balanced operating point where the marginal utility of scaling is nearly equilibrated, providing the most favorable compromise for mobile deployment.

\subsubsection{Summary}
Collectively, these studies validate our pipeline: architectural design enables efficiency and the chosen model scale optimally balances both objectives. Any deviation yields a strictly inferior solution, demonstrating the necessity of each design choice.

\section{Conclusion and Future Work}
\label{sec: conclusion}

In this paper, we have addressed the critical challenge of deploying powerful MOBA game AI agents on resource-constrained mobile devices.
we propose a Pareto-guided distillation approach and design an efficient, featherweight student architecture tailored for MOBA games and mobile deployment. Our approach significantly increases inference speed and energy efficiency while maintaining competitive performance in practice.

Although our experiments focus on HoK, the proposed methodology establishes a general pipeline for efficient policy compression and deployment in real-time, resource-limited settings. The demonstrated ability to retain strategic competence under efficiency constraints represents a meaningful step toward bridging the gap between large-scale game AI systems and practical mobile deployment.

In future work, we plan to extend our study toward more general mobile game environments and real-time decision-making tasks, exploring adaptive distillation strategies and hardware-aware co-design to further enhance scalability and robustness.



\begin{acks}
We would like to express our sincere gratitude to TiMi L1 Studio for their valuable support in providing services through the Tencent AI Arena platform(https://tencentarena.com).
\end{acks}



\bibliographystyle{ACM-Reference-Format} 
\bibliography{references}

\appendix
\section{HoK Environment}
\label{app sec: hok env}
\subsection{Observation Space}
\label{app ssec: observation space}

This appendix provides a comprehensive description of the observation space representation used in our Honor of Kings 3v3 environment. The observation space is architected as a high-dimensional, multi-modal vector representation that encodes the complete game state at each time step, enabling informed decision-making by the agent.

The observation vector has a total dimensionality of 13,758, systematically decomposed into three hero-specific components of 4,586 dimensions each. This tripartite structure allows parallel processing of each hero's perceptual information while maintaining a unified game state representation. The observation space captures seven distinct categories of game information, each serving specific functional roles in the decision-making process:

\begin{itemize}
    \item \textbf{Image-like Features} (1,734 dimensions): Represent spatial information through six semantic channels (obstacles, terrain features, grass visibility, etc.) structured in a 17×17 grid format, providing a spatial awareness basis for navigation and tactical positioning.
    
    \item \textbf{Hero Information} (1,506 dimensions): Encodes the state of all six heroes in the match (both allies and enemies), including identifiers, health points, skill cooldowns, buff/debuff statuses, and combat-relevant attributes.
    
    \item \textbf{Current Hero Private Features} (44 dimensions): Captures hero-specific information only available to the controlled unit, such as enemy presence within attack range, target acquisition status, and local tactical context.
    
    \item \textbf{Soldier Status} (500 dimensions): Tracks the state of 20 creeps (both allied and enemy), including unit type, health, position, combat state, and target information.
    
    \item \textbf{Turret Status} (174 dimensions): Monitors the condition of 6 turrets throughout the map, containing turret type, health points, defensive status, and positional information.
    
    \item \textbf{Monster Status} (560 dimensions): Encodes the state of 20 neutral monsters, including monster type, health, position, respawn timers, and combat engagement status.
    
    \item \textbf{Global Game Information} (68 dimensions): Provides macroscopic game state information, including team gold economies, kill counts, surviving turrets, and objective-based metrics that inform strategic decision-making.
\end{itemize}

As detailed in Table~\ref{tab: obs space}, this comprehensive feature organization enables agents to maintain a holistic perception of the game state, balancing detailed unit-level information with strategic global context. The structured organization facilitates efficient feature extraction and processing while ensuring all relevant game elements are represented for optimal decision-making.

The multi-modal nature of this observation space—incorporating spatial, temporal, categorical, and numerical data types—poses significant challenges for representation learning, but provides the necessary foundation for sophisticated strategic behavior in the complex MOBA environment.

\begin{table}[h]
    \centering
    \caption{Observation space composition in HoK 3v3 mode: feature categories, semantic descriptions, and dimensions.}
    \label{tab: obs space}
    \begin{tabular*}{\linewidth}{@{}p{2cm}p{4cm}p{2cm}@{}}
        \toprule 
        \textbf{Categories of Features} & \multicolumn{1}{c}{ \textbf{Descriptions} } & \textbf{Dimensions} \\
        \midrule
        Image-like Features &  Image-like features, including 6 channels such as obstacle channel and grass channel.& $6*17*17$ \\\hline 
        Heroes & From vision of the current player, state information of 6 heroes from both sides, i.e., hero ID, HP, etc. & $6 * 251$ \\\hline
        Current Hero & Private features of the current hero, i.e., whether the enemy hero is within the attack range of the current hero.& $44$ \\\hline
        Soldiers & The state of 20 Creep of allies and enemies: types, HP, positions, etc. & $20 * 25$ \\\hline
        Turrets & The state of 6 Turrets: types, HP, positions, etc. & $6 * 29$ \\\hline
        Monsters & The state of 20 Monsters: types, HP, positions, etc. & $20 * 28$ \\\hline
        Whole Info & Golds of allies and enemies; kills, surviving turrets, etc. & 68 \\
        \bottomrule
    \end{tabular*}
\end{table}

\subsection{Action Space}
\begin{figure*}[h]
    \centering
    \includegraphics[width=1\linewidth]{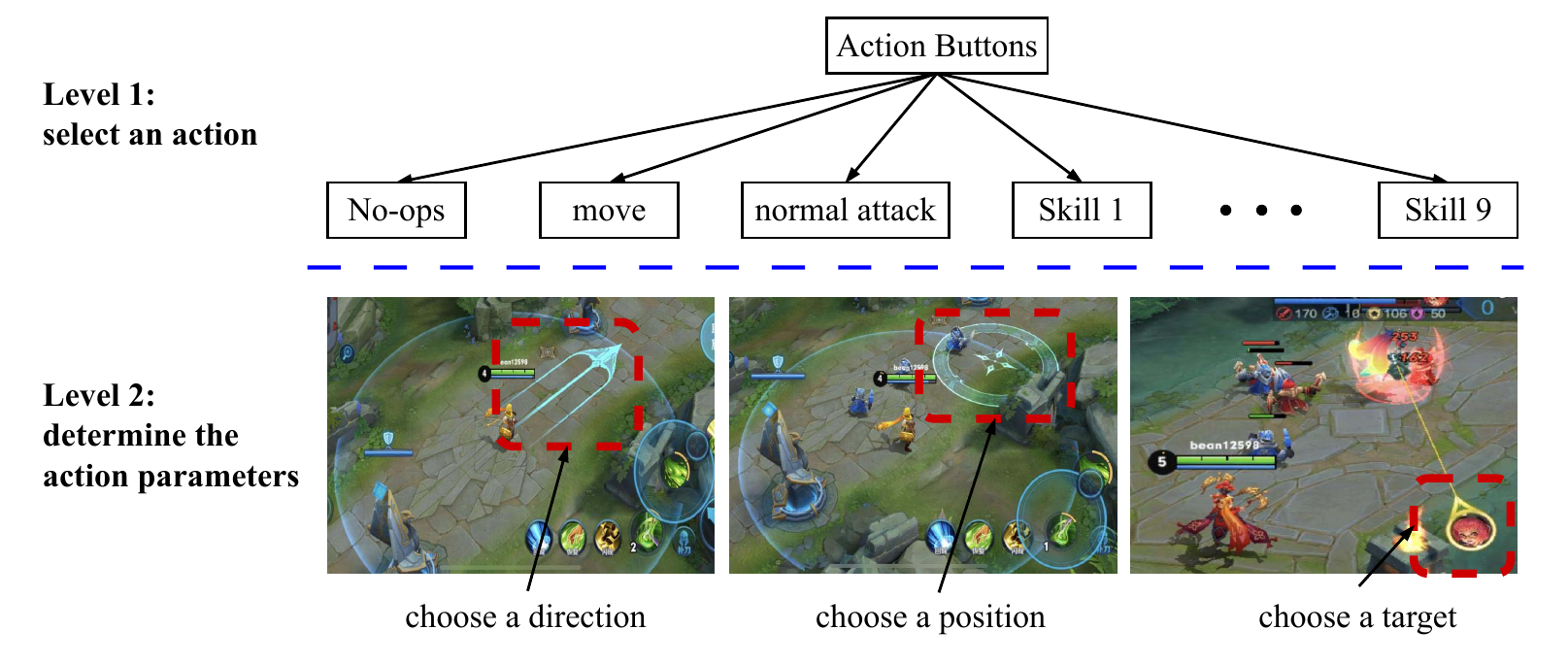}
    \caption{Schematic diagram of hierarchical action space in HoK 3v3 mode.}
    \label{fig: two level concept}
\end{figure*}

The action space employs a two-level hierarchical structure to enable precise micromanagement of heroes, as outlined in Table~\ref{tab: action space} and detailed in Table~\ref{tab: detail action space}, and the hierarchical concept diagram of the action space is shown in Figure~\ref{fig: two level concept}. The first level (select an action, row 1 in Table~\ref{tab: detail action space}) governs the behavior type through 13 discrete button actions, comprising two no-operation (no-op) commands, a Move action, a Normal Attack, and nine hero-specific skills. The second level (determine the action parameters, row 2 - 4 in Table~\ref{tab: detail action space}) specifies the action parameters through three distinct subcategories: 
(1) Direction Control, offering 25 discrete movement angles; 
(2) Position Control, allowing skill adjustments across a 42×42 grid in the x-z plane for precise targeting; and 
(3) Target Selection, which supports seven target types (e.g., enemy heroes, allies, monsters, minions, and turrets) with up to 39 possible targets. 

\begin{table*}[h]
\caption{Action space description.}
\label{tab: action space}
\centering
\begin{tabular*}{\linewidth}{@{}@{\extracolsep{\fill}}lll@{}}
    \toprule  
    \textbf{Categories} & \textbf{Level} & \textbf{Specification}\\
    \midrule
    Action Selection & 1 & 13 discrete actions: 2 no-ops, move, normal attack, and 9 skills \\
    Direction Control & 2 & 25 discrete directions for action orientation \\
    Position Control & 2 & 42×42 possible skill casting positions (x-z plane) \\
    Target Selection & 2 & 7 target types with 39 possible targets \\
    \bottomrule
\end{tabular*}
\end{table*}

\begin{table*}[htbp]
    \centering
    \caption{Detailed specification of action space.}
    \label{tab: detail action space}
    \begin{tabular*}{\linewidth}{@{}@{\extracolsep{\fill}}lllc@{}}
        \toprule
        \textbf{Action Categories} & \textbf{Sub Action} & \textbf{Description} & \textbf{Dimension} \\
        \midrule
        \multirow{13}{*}{Action Button} 
            & None & Inactive state & 1  \\
            & None & Inactive state & 1  \\
            & Move & Move hero & 1  \\
            & Normal Attack & Cast normal attack & 1  \\
            & Skill 1 & Cast skill 1 & 1  \\
            & Skill 2 & Cast skill 2 & 1  \\
            & Skill 3 & Cast skill 3 & 1  \\
            & Skill 4 & Cast skill 4 (hero-specific) & 1  \\
            & Chosen Skill & Cast selected skill & 1  \\
            & Recall & Return to base & 1  \\
            & Equipment Skill & Activate equipment skill & 1  \\
            & Heal Skill & Cast heal skill & 1  \\
            & Friend Skill & Cast ally skill (hero-specific) & 1  \\
        \midrule
        Move & Movement Directions & \makecell[l]{Movement direction \\ (25 discrete angles)} & 25 \\
        
        \midrule
        \multirow{2}{*}{Skill Offset} 
            & X-axis & Skill horizontal adjustment & 42 \\
            & Z-axis & Skill vertical adjustment & 42 \\
        \midrule
        \multirow{7}{*}{Target}
            & None & No target & 1  \\
            & Enemy Heroes & 3 enemy heroes & 3  \\
            & Friend Heroes & 3 allied heroes & 3  \\
            & Self & Own hero & 1  \\
            & Monster & 20 jungle monsters & 20 \\
            & Soldier & 10 closest minions & 10 \\
            & Turret & Closest turret & 1  \\
        \bottomrule
    \end{tabular*}
\end{table*}

To ensure action validity and contextual relevance, HoK 3v3 mode integrates legal action masks and sub-action masks (explained in~\ref{app ssec: legal action mask} and ~\ref{app ssec: sub action mask} respectively), building on established methodologies from prior research~\cite{ye2020mastering, ye2020towards}. This structured approach balances granular control with computational efficiency, critical for complex real-time strategy environments.

\subsection{Legal Action Mask Mechanism}
\label{app ssec: legal action mask}
during each time step of an episode, every hero must select an action from a constrained set of valid choices. To enforce this restriction, a legal action mask is applied, dynamically filtering out invalid actions based on contextual rules. For action types such as Button, Move, and Skill, the mask’s dimensionality aligns directly with the action space—ensuring only permissible options (e.g., available skills or movement directions) are selectable. However, the Target action type exhibits a dependency on the chosen Button: its legality is contingent on the button selected. For instance, certain skills may require targeting allies, while others are limited to enemies or neutral units. Given the 13 possible Button actions and 7 Target categories (e.g., heroes, minions, turrets), the Target mask adopts a 13×7 matrix structure, where each button-target pair is validated against game-state rules. This hierarchical masking system ensures both flexibility and adherence to gameplay constraints, preventing nonsensical or rule-violating actions while maintaining strategic depth.

\subsection{Sub-Action Mask Mechanism}
\label{app ssec: sub action mask}
The sub-action masking process dynamically filters out incompatible actions that cannot be executed concurrently with the currently selected primary action, thereby preserving only contextually permissible options. This hierarchical constraint system ensures action validity and operational coherence within the game environment. Two representative examples demonstrate its implementation:

\textbf{Example 1 (Movement Control):}
When the Button-Move action is activated, the masking mechanism retains only the directional control sub-actions (Move Dir). These sub-actions govern movement trajectories through discrete directional vectors, while automatically suppressing irrelevant parameters like skill targets or positional offsets.

\textbf{Example 2 (Combat Execution):}
Selection of the Button-Normal Attack triggers a complementary masking effect, where only the Target sub-action persists. This enforces a valid attack sequence by:
(1) Maintaining target selection parameters (e.g., enemy units within attack range);
(2) Eliminating mutually exclusive options (e.g., movement directions or skill cast positions).

The same masking logic extends to all actions, though the exact sub-action masks differ based on each hero's characteristics and equipped items.

\begin{figure}[h]
    \centering
    \includegraphics[width=0.5\linewidth]{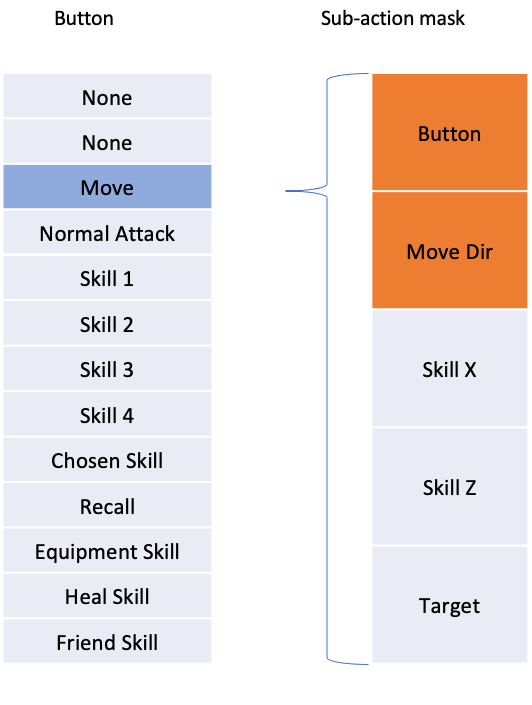}
    \caption{Schematic diagram of Sub Action mask after selecting the Button-Move action}
    \label{fig: sub action mask}
\end{figure}

\section{Heroes Lineup for Evaluation}
\label{app sec: hero compositions}

This appendix details the hero compositions utilized for agent evaluation in our study. To ensure a robust and comprehensive assessment of agent performance, we employ a systematically generated set of eight distinct team compositions. These compositions are derived from a pool of six popular and strategically diverse heroes frequently encountered in high-level Honor of Kings gameplay.

The selection pool consists of three positional pairs, with each position offering two hero choices (show in Table~\ref{tab:hero_compositions}):
\begin{itemize}
    \item \textbf{Position 1 (Frontline/Bruiser):} Zhong Wuyan (control/tank) or Zhao Yun (mobile assassin)
    \item \textbf{Position 2 (Marksman/Damage Dealer):} Sun Shangxiang (sustained damage) or Li Yuanfang (burst damage)
    \item \textbf{Position 3 (Mage/Support):} Zhuge Liang (ranged burst) or Diao Chan (control/mobility)
\end{itemize}

The complete set of eight team compositions, presented in Table~\ref{tab:team_compositions}, encompasses all possible combinations from this selection pool. This systematic approach ensures coverage of diverse team synergies and strategic archetypes, including: burst damage compositions, sustained damage compositions, balanced teams, and control-oriented teams.

This methodological choice mitigates evaluation bias that could arise from testing agents on a single, potentially favorable composition. By evaluating performance across this varied set of scenarios, we obtain a more reliable and generalizable assessment of an agent's capability to handle the strategic diversity inherent in competitive MOBA environments.

\begin{table}[h]
\centering
\caption{Hero composition for evaluation}
\label{tab:hero_compositions}
\begin{tabular}{|c|l|l|l|}
\hline
Position & \multicolumn{1}{c|}{Hero Option 1} & \multicolumn{1}{c|}{Hero Option 2} \\
\hline
1 & Zhong Wuyan & Zhao Yun \\
2 & Sun Shangxiang & Li Yuanfang \\
3 & Zhuge Liang & Diao Chan \\
\hline
\end{tabular}
\end{table}

\begin{table}[h]
\centering
\caption{The eight team compositions used for evaluation}
\label{tab:team_compositions}
\begin{tabular}{|c|l|}
\hline
Composition ID & \multicolumn{1}{c|}{Hero Composition} \\
\hline
1 & Zhao Yun, Li Yuanfang, Zhuge Liang \\
2 & Zhao Yun, Li Yuanfang, Diao Chan \\
3 & Zhao Yun, Sun Shangxiang, Diao Chan \\
4 & Zhao Yun, Sun Shangxiang, Zhuge Liang \\
5 & Zhong Wuyan, Li Yuanfang, Zhuge Liang \\
6 & Zhong Wuyan, Li Yuanfang, Diao Chan \\
7 & Zhong Wuyan, Sun Shangxiang, Diao Chan \\
8 & Zhong Wuyan, Sun Shangxiang, Zhuge Liang \\
\hline
\end{tabular}
\end{table}

\section{Architecture of the Teacher Model}
\label{app sec: teacher architecture}

\begin{figure*}[h]
    \centering
    \includegraphics[width=\linewidth]{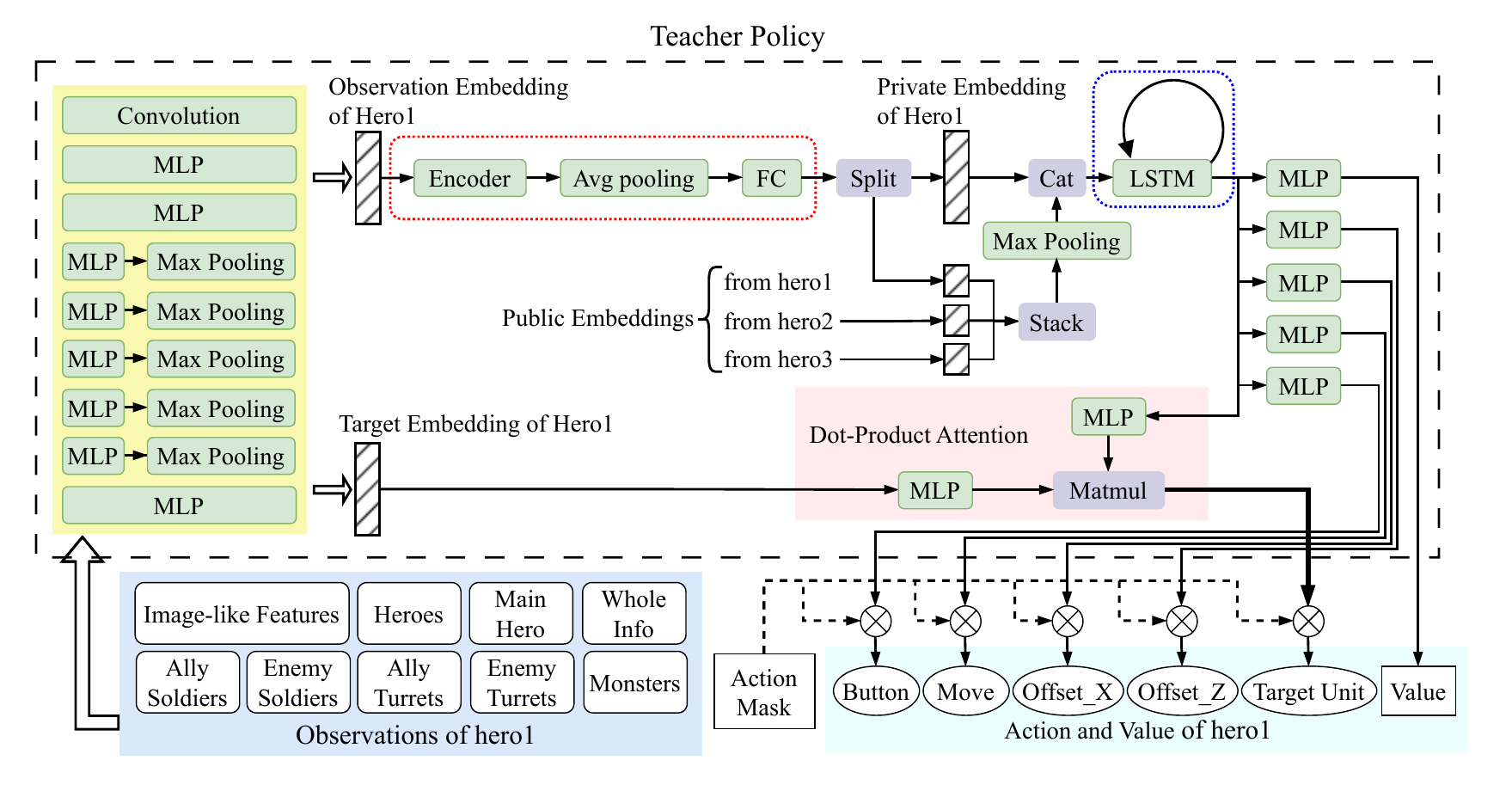}
    \caption{Architecture overview of the teacher model, illustrated using hero1's data flow. Key procedures include: (1) Input observations (light blue background) containing 9 feature categories; (2) The observation embedding and target embedding are obtained by extracting features (detailed in yellow background rectangle) through the convolutional layer and MLP layers; (3) Using Encoder (with avg pooling and FC layers, red dashed box) for spatial encoding (4) Using LSTM for temporal encoding (blue dashed box); (5) dot-product attention (light red background) using dot-product operations between hero1's target embedding and public embeddings from other 2 heroes; (6) Output predictions (cyan background) generating masked actions and state value. Additionally, arrows indicate information flow between neural modules (light green rectangles), operators (light purple rectangles), and intermediate embeddings (tensor, striped filled rectangles). The complete teacher policy is demarcated by the black dashed boundary. More detailed structural information and dimensional parameters about the teacher model are presented in the Figure~\ref{fig: detail teacher net}.}
    \label{fig: teacher net}
\end{figure*}

This appendix provides a comprehensive description of the teacher model architecture that serves as the foundation for our knowledge distillation process. The teacher model, building upon prior work in hierarchical decision-making~\cite{ye2020supervised, ye2020mastering, ye2020towards}, is designed to handle the complex dynamics of the HoK 3v3 environment. As depicted in Figure~\ref{fig: teacher net}, the model processes observations from all three heroes and outputs corresponding action predictions and state values for each hero.

\subsection{Architectural Components}

The teacher model employs a sophisticated multi-component architecture that directly informs the design of our featherweight student model:

\textbf{Feature Extraction Module}: The input pipeline begins with structured processing of nine observation categories (detailed in Table~\ref{tab: obs space}). This module employs specialized convolutional layers for spatial inputs and multi-layer perceptrons (MLPs) for scalar features, extracting relevant features from each modality. The processed features are concatenated to form a comprehensive observation embedding for each hero.

\textbf{Feature Fusion Module}: Spatial encoding is performed through a transformer encoder architecture incorporating average pooling for dimensionality reduction and fully-connected layers for feature transformation. This module effectively fuses multi-modal features into a unified representation.

\textbf{Triplet Fusion Gate}: The model facilitates inter-hero communication by splitting embeddings into private and public components. The public components undergo max-pooling across all three heroes, enabling effective team coordination and strategic alignment.

\textbf{Temporal Fusion Module}: Temporal dynamics are captured through an LSTM network processing sequences of 16 consecutive frames. This module is crucial for inferring latent game states and capturing motion patterns under partial observability.

\textbf{Prediction Heads}: The hierarchical action space is addressed through specialized prediction heads that sequentially determine action types, select targets using dot-product attention mechanisms, and estimate state values. Notably, target selection focuses exclusively on enemy features for combat efficiency.

\subsection{Architecture Details and Dimensional Parameters\label{app sec: teacher architecture detail}}
The teacher model processes observations from three heroes through a structured input pipeline that handles nine distinct observation categories, including image-like features (6×17×17), hero attributes (6×251), current hero status (44), ally/enemy soldiers (10×25 each), ally/enemy turrets (3×29 each), monsters (20×28), and global state information (68). Feature extraction begins with specialized layers: convolutional layers (e.g., 6→18 channels, kernel size 5, stride 1, padding 2) process spatial inputs, while fully connected (FC) layers (e.g., 251→1024, 44→256) handle non-spatial features, all followed by ReLU activation. These extracted features are concatenated into a unified observation embedding (224-dimensional) for each hero, which is then processed by a 3-layer multi-head attention encoder (4 heads, hidden size 224). The encoder's output undergoes average pooling (e.g., 19×1) and is projected to a higher dimension via an FC layer (224→1024).  

The model further divides the embeddings into private (hero-specific, 768-dimensional) and public (max-pooled across all three heroes, 256-dimensional) components. These are concatenated and fed into a single-layer LSTM (hidden size 1024, non-bidirectional, dropout disabled) to capture temporal dependencies across 16-frame sequences. For decision-making, the architecture employs five specialized prediction heads: FC layers (e.g., 224→25 for movement actions, 224→42 for spatial offsets) and a dot-product attention mechanism for target selection (using enemy-focused embeddings). To ensure training efficiency, the model incorporates action masking (161-dimensional) to restrict invalid choices and a value prediction head (224→1) for state evaluation.  

\begin{figure*}[ht]
    \centering
    \includegraphics[width=\linewidth]{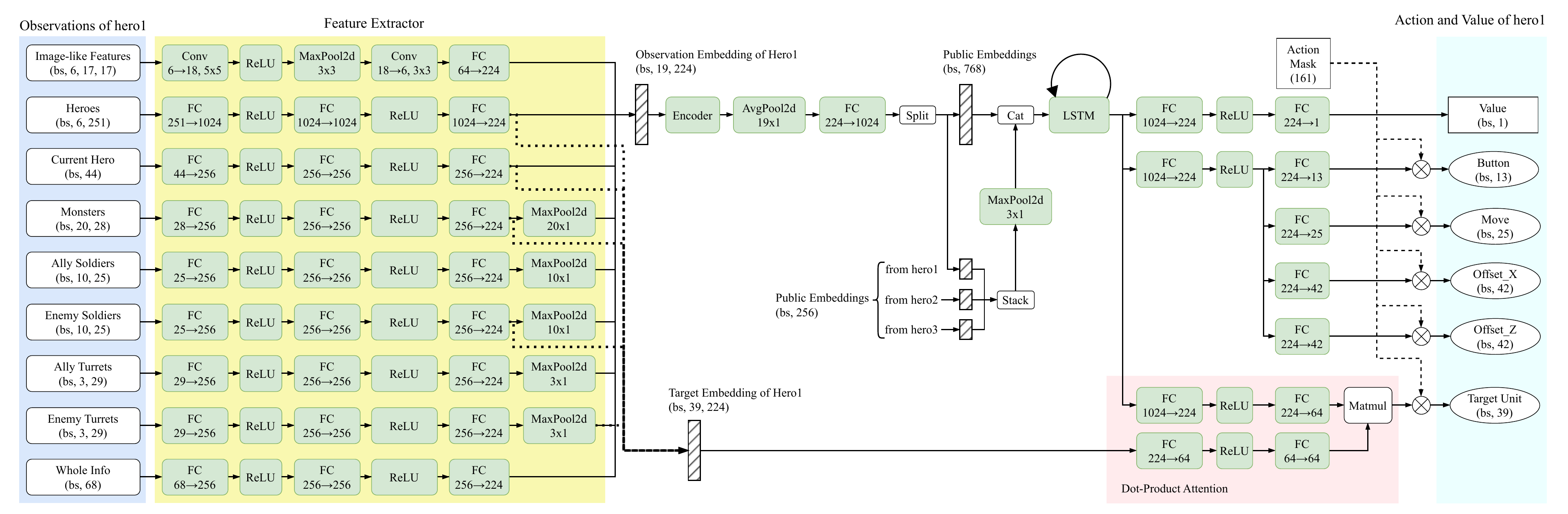}
    \caption{The detailed architecture of the teacher model. Note that "bs" in the figure represents the batch size, the numbers below the layer represent the key parameters related to this layer, and the dimension information is marked in parentheses below the tensor}
    \label{fig: detail teacher net}
\end{figure*}

Detailed structural information and dimensional parameters of the teacher model are illustrated in Figure~\ref{fig: detail teacher net}, where "bs" denotes batch size.  

\section{Hyper-parameters}
\label{app sec: hyper_parameters}
This section presents the hyper-parameter settings for distillation training, as detailed in Table~\ref{tab:hyper_parameter}.

\begin{table}[h]
    \caption{Hyper-parameter settings for distillation training.}
    \label{tab:hyper_parameter}
    \centering
    \begin{tabular*}{\linewidth}{ccp{4cm}}
    \toprule
    \textbf{Hyper-Parameters}&\textbf{Settings}&\textbf{Descriptions}\\
    \midrule
    Random seed & 56 & - \\
    Optimizer & Adam & - \\
    Learning rate & 0.0002 & Adam learning rate \\
    $\phi_1$ & 0.9 & Adam decay rate 1 \\
    $\phi_2$ & 0.999 & Adam decay rate 2 \\
    Adam-eps & $1 \times 10^{-8}$ & Adam epsilon \\
    Batch Size& 512 & - \\
    T-max & $1 \times 10^6$ & Number of training steps \\
    $\tau$ & 4 & Temperature parameter for distillation \\
    \bottomrule
    \end{tabular*}
\end{table}

\section{Search Space}
\label{app sec: search space}

This appendix provides a comprehensive description of the architectural search space employed in our methodology for generating diverse candidate student models. The search space is systematically designed to explore the micro-architectural variations within the macro-architecture template defined in Section~\ref{ssec: architecture design}, enabling a thorough investigation of the performance-efficiency trade-off landscape.

\subsection{Search Space Structure}

The search space is decomposed into seven fundamental building blocks that correspond to the core components of our student architecture template. Each block represents a distinct functional unit within the neural network architecture and is characterized by two primary hyperparameters:
\begin{itemize}
    \item \textbf{Layer Count ($layers$)}: The number of layers within the block, defining the architectural depth
    \item \textbf{Feature Dimensions ($n_i$)}: The width of each layer, specified for the $i$-th layer within the block
\end{itemize}

These hyperparameters are carefully designed to capture the essential degrees of freedom in the model architecture while maintaining functional integrity and computational feasibility.

\subsection{Block-Wise Parameterization}

The complete parameterization of each building block is detailed in Table~\ref{tab: search_space}, which specifies the allowable ranges for each hyperparameter:

\begin{itemize}
    \item \textbf{Token Dimension Block}: Controls the embedding dimension for tokenized inputs, with $n_1$ ranging from 32 to 224 features
    \item \textbf{Attention Dimension Block}: Governs the dimensionality of attention mechanisms, with $n_1$ varying between 32 and 128 features
    \item \textbf{Role Feature Transformation Block}: Handles role-specific feature processing, with $n_1$ and $n_2$ ranging from 32 to 256 features and layer depth of 1-2
    \item \textbf{Image Feature Transformation Block}: Processes visual features, supporting $n_1$ up to 1024 and $n_2$ up to 512 features with 1-2 layers
    \item \textbf{Concatenated Feature Transformation Block}: Manages fused multi-modal features, offering the largest capacity with $n_1$ up to 4096, $n_2$ up to 2048, and predefined $n_3^+$ dimensions, spanning 1-5 layers
    \item \textbf{Communication Feature Transformation Block}: Facilitates inter-hero communication, with balanced dimensions ($n_1$ and $n_2$ up to 1024, $n_3^+$ 32-512) and 0-3 layers
    \item \textbf{Action Prediction Block}: Handles final action distribution generation, with $n_1$ constrained to 32-256 features
\end{itemize}

\subsection{Constrained Sampling Strategy}

To ensure computational tractability and representative coverage of the design space, we implement a structured sampling approach with two key constraints:

\textbf{Global FLOPs Budget}: Candidate architectures are constrained to a computational budget ranging from 1\% to 20\% of the teacher model's FLOPs (673.44 MFLOPs), ensuring all generated models are suitable for mobile deployment.

\textbf{Stratified Sampling}: The FLOPs range is partitioned into 20 discrete intervals (each representing 1\% of the teacher's FLOPs), with uniform sampling within each interval. This approach guarantees balanced representation across the entire performance-efficiency spectrum and prevents over-sampling in computationally favorable regions.

This systematic exploration strategy generates a diverse population of candidate models that effectively populate the architectural design space, enabling comprehensive analysis of the Pareto-optimal frontier and facilitating the identification of optimal performance-efficiency trade-offs.

\begin{table*}[h]
\caption{Search space: $n_3^+$ records the upper limit of the dimension of each layer of each block starting from the third layer, and the lower limit is 32.}
\label{tab: search_space}
\centering
\begin{tabular*}{\linewidth}{@{}@{\extracolsep{\fill}}lrrrr@{}}
    \toprule  
    \textbf{BlockType} & $n_1$&$n_2$&$n_3^+$&$layers$\\
    \midrule
    token\_dim & 32$\sim$224& - & - & - \\
    attention\_dim & 32$\sim$128& - & - & - \\
    role\_feat\_trans & 32$\sim$256& 32$\sim$256& -&1$\sim$2\\
    img\_feat\_trans & 32$\sim$1024& 32$\sim$512& -&1$\sim$2\\
    concat\_feat\_trans & 32$\sim$4096& 32$\sim$2048& [1024,1024,512]&1$\sim$5\\
    communicate\_feat\_trans & 32$\sim$1024& 32$\sim$1024& 32$\sim$512&0$\sim$3\\
    action\_fc & 32$\sim$256& -& -&-\\
    \bottomrule
\end{tabular*}
\end{table*}

\begin{table*}[ht]
\caption{The final result of the competition.}
\label{tab: result of the match}
\centering
\small
\begin{tabular*}{\linewidth}{@{}@{\extracolsep{\fill}}lrrrrrr@{}}
    \toprule  
    \textbf{Ranking} & \textbf{Win Rate} & \textbf{Time(s)} & \textbf{Memory(MB)} & \textbf{Energy(mAh)} & \textbf{Size(MB)} & \textbf{Score} \\
    \midrule
    1st & 38.67\% ($\pm$1.25\%)  & 2.33 ($\pm$0.03)  & 156.71 ($\pm$0.19)  & 0.56 ($\pm$0.01)  & 11.59  & 87.49\\
    2nd & 36.5\% ($\pm$0.41\%)  & 8.35 ($\pm$0.12)  & 162.59 ($\pm$0.94)  & 1.12 ($\pm$0.02)  & 12.87  & 75.54\\
    3rd & 30.83\% ($\pm$2.39\%)  & 1.84 ($\pm$0.01)  & 155.77 ($\pm$3)  & 0.34 ($\pm$0.01)  & 5.32  & 72.56\\
    4th & 20\% ($\pm$2.16\%)  & 4.7 ($\pm$0.11)  & 165.25 ($\pm$0.43)  & 0.73 ($\pm$0.01)  & 4.59  & 46.18\\
    5th & 17.83\% ($\pm$0.47\%)  & 4.29 ($\pm$0.08)  & 165.05 ($\pm$0.86)  & 0.46 ($\pm$0.02)  & 1.52  & 42.74\\
    \bottomrule
\end{tabular*}
\end{table*}

\section{Competition Overview}
\label{app sec: competition}

This appendix provides a detailed description of the competition associated with the main task. The objective of the competition was to develop a lightweight policy model for the mobile 3v3 mode of HoK that could be efficiently deployed on resource-constrained mobile devices. Participants were provided with a high-performance, computationally intensive "teacher" agent and were challenged to produce a distilled or optimized "student" agent that maintained the original model's strategic decision-making quality while significantly improving inference efficiency.

\subsection{Evaluation Framework}

The primary goal was to balance performance preservation with operational efficiency. A comprehensive evaluation framework was designed to assess submissions based on the following four key metrics, which represent critical constraints for mobile deployment:

\begin{enumerate}
    \item \textbf{Latency:} Average inference time (in seconds) measured on the target mobile device.
    \item \textbf{Memory Usage:} Average runtime memory overhead (in MB) during model execution.
    \item \textbf{Battery Consumption:} Average energy drain (in mAh) incurred during the inference process.
    \item \textbf{Model Size:} Final storage footprint (in MB) of the deployed model on the mobile device.
\end{enumerate}

\begin{table}[ht]
\caption{Strong and weak reference indicators for the scoring formula.}
\label{tab: reference of match}
\centering
\small
\begin{tabular*}{\linewidth}{@{}@{\extracolsep{\fill}}lrr@{}}
    \toprule
    \textbf{Metric} & \textbf{Strong Reference} & \textbf{Weak Reference}\\
    \midrule
    Latency (s) & 15 & 315 \\
    Memory (MB) & 160 & 230 \\
    Energy (mAh) & 2 & 75 \\
    Size (MB) & 7 & 80 \\
    \bottomrule
\end{tabular*}
\end{table}

\subsection{Scoring Methodology}

Each of the four efficiency metrics was converted into an individual sub-score. Lower measured values indicate better performance for all metrics. To standardize scoring across metrics with different units and scales, a logarithmic scaling function was used, anchored by predefined strong and weak reference values derived from the performance distribution of baseline models. The reference values are listed in Table~\ref{tab: reference of match}.

The sub-score for a given metric was calculated as follows:
\begin{equation}
\text{Sub-score} = \min\left(\frac{\log(W) - \log(V)}{\log(W)) - \log(S)} \times 60 + 20,\ 100\right),
\end{equation}
where $W$ is weak reference, $S$ is strong reference and $V$ is measured value.

This formula maps performance onto a scale from 20 to 100, capped at 100, where a score of 100 indicates performance equal to or better than the strong reference.

The raw composite score was computed as a weighted sum of the non-negative sub-scores:
\begin{equation}
\begin{aligned}
\text{Raw Score} = &\max(\text{Latency score}, 0) \times 0.4 \\
&+ \max(\text{Memory score}, 0) \times 0.2 \\
&+ \max(\text{Energy score}, 0) \times 0.2 \\
&+ \max(\text{Size score}, 0) \times 0.2
\end{aligned}
\end{equation}
Latency was assigned the highest weight (0.4) due to its critical importance for real-time interaction.

Finally, to ensure that efficiency gains did not come at the cost of strategic capability, the raw score was scaled by the model's effectiveness. The win rate of the submitted agent against the original teacher model was measured, and the final score was adjusted accordingly:
\begin{equation*}
\text{Final Score} = \min\left(\frac{\text{Measured Win Rate}}{\text{Target Win Rate}},\ 1\right) \times \text{Raw Score}
\end{equation*}
A target win rate of 40\% was set, indicating that the original policy's decision-making quality had been satisfactorily preserved.

\subsection{Competition Results}

The final results for the top five competitors are presented in Table~\ref{tab: result of the match}. The winning entry achieved the highest final score by successfully balancing a near-target win rate (38.67\%) with strong performance across all efficiency metrics, particularly latency and energy consumption.

\subsection{Performance Analysis of Top Competitors\label{sec: student performance}}
\begin{figure}[h]
    \centering
    \includegraphics[width=1.0\linewidth]{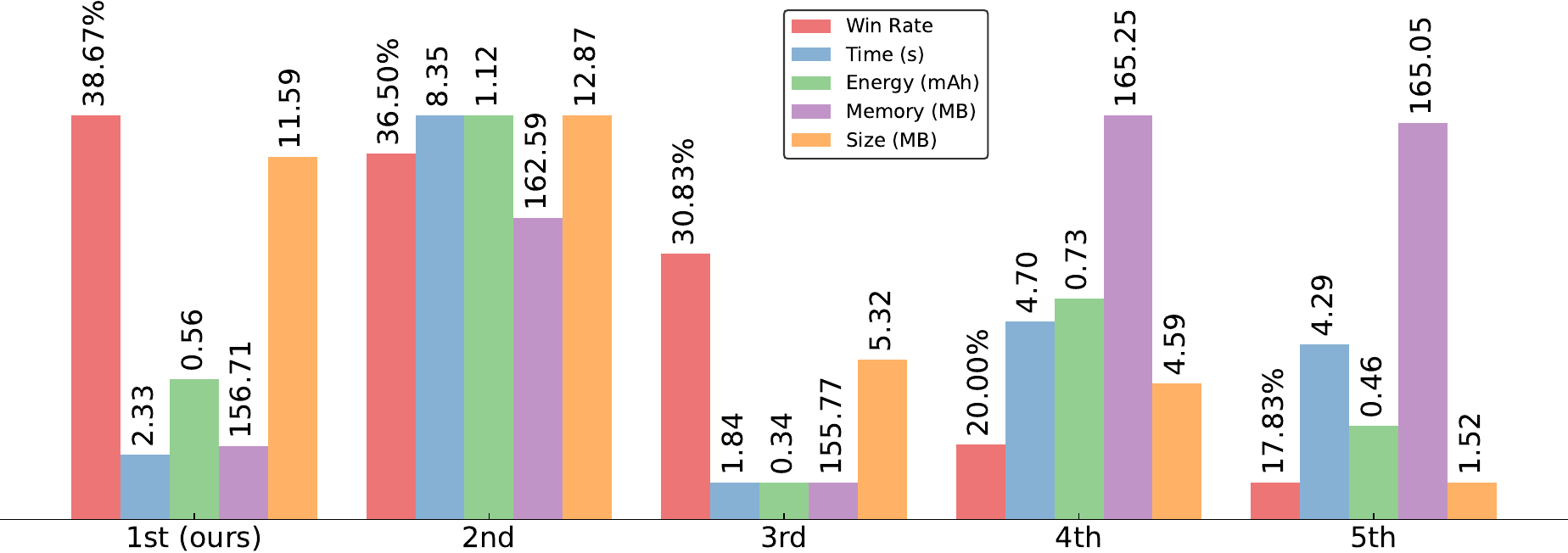}
    \caption{Radar chart illustrating the normalized performance of the top five competition entries across the five primary evaluation metrics (detailed values are provided in Table~\ref{tab: result of the match}).}
    \label{fig: top 5 competitor}
\end{figure}
The results highlight the inherent trade-off between model efficiency and strategic performance. The 1st-place model excelled by achieving the highest win rate and leading in inference speed and energy efficiency, resulting in the top final score. Notably, the 3rd-place model achieved the best latency and model size, and the second-best memory and energy usage, but was penalized by a lower win rate, demonstrating the critical importance of the performance preservation clause in the scoring system. A comparative visualization of the top five entries across the key metrics is provided in Figure~\ref{fig: top 5 competitor}.

The competition successfully fostered the development of models capable of high-frequency, energy-efficient inference on mobile devices while maintaining competitive gameplay performance against a strong baseline. The winning entry, in particular, demonstrated a superior balance, making it suitable for real-time deployment scenarios.

\end{document}